\documentclass[11pt,a4paper]{article}
\usepackage[hyperref]{acl2020}
\usepackage{times}
\usepackage{latexsym}

\usepackage{microtype}

\usepackage[utf8]{inputenc}
\usepackage{makecell} 
\usepackage{multirow} 
\usepackage{multicol}
\usepackage{makecell}
\usepackage{tabularx}
\usepackage{color}
\usepackage{xcolor}
\usepackage{listings}
\usepackage{subcaption}
\usepackage{url}
\usepackage{amsmath}
\usepackage{booktabs}
\usepackage{adjustbox}

\newcommand{\ignore}[1]{}

\aclfinalcopy

\title{Multi-task Retrieval for Knowledge-Intensive Tasks}

\makeatletter
\newcommand{\printfnsymbol}[1]{%
  \textsuperscript{\@fnsymbol{#1}}%
}
\makeatother

\author{
    Jean Maillard\thanks{\hspace{.04in} Equal Contribution.} \quad Vladimir Karpukhin\printfnsymbol{1} \quad Fabio Petroni \\
    {\bf Wen-tau Yih} \quad {\bf Barlas O{\u{g}}uz} \quad {\bf Veselin Stoyanov} \quad {\bf Gargi Ghosh} \\
    Facebook AI \\
    \texttt{\{jeanm,vladk,fabiopetroni,scottyih,barlaso,ves,gghosh\}@fb.com} \\  
}

\begin{document}

\maketitle

\begin{abstract}

Retrieving relevant contexts from a large corpus is a crucial step for tasks such as open-domain question answering and fact checking. Although neural retrieval outperforms traditional methods like tf-idf and BM25, its performance degrades considerably when applied to out-of-domain data. 
Driven by the question of whether a neural retrieval model can be \emph{universal} and perform robustly on a wide variety of problems, we propose a multi-task trained model.
Our approach not only outperforms previous methods in the few-shot setting, but also rivals specialised neural retrievers, even when in-domain training data is abundant.
With the help of our retriever, we improve existing models for downstream tasks and closely match or improve the state of the art on multiple benchmarks.

\end{abstract}

\ignore{
	• Introduction
	• Background
		○ Retrieval methods -- BM25/TFIDF, Dense Retrieval
		○ NLP applications -- KILT
	• Approach -- universal retriever
	• Experiments
	• (Other) Related Work
	• Conclusions
}

\section{Introduction}
\label{sec:intro}

\ignore{

	• Retrieval is important
	• Dense representation can replace TFIDF/BM25 (now)
		○ Vanilla pre-trained model underperforms TFIDF/BM25 with a big gap
		○ Domain-specific fine-tuning is still required
	• The problem studied in this work: Making dense representation more "robust"
		○ Without fine-tuning, still performs reasonably
		○ With some in-domain labeled data, can be tuned future to improve the performance
	• Solution
		○ Multi-task learning that is trained using multiple "domains" of tasks
			§ The goal is not to improve one particular single-task performance, but instead making sure the model performs reasonably well on out-of-domain tasks/data
		○ More here?
	• Contributions
		○ A more robust version of dense representation
		○ Models are readily usable
		○ Good performance on KILT
}

\emph{Knowledge-intensive tasks} is the common designation for a class of real-world NLP problems which, because of their nature, require large amounts of knowledge about the world~\citep{kilt}. 
For example, \emph{open-domain question answering} requires producing answers to general factoid questions; \emph{fact checking} involves determining the veracity of claims based on a database of trusted evidence.
Practical solutions to these tasks usually involve an efficient \emph{retrieval} component that, given an input query, selects a limited subset of relevant information from a large knowledge source. Sophisticated downstream models then consider the input only in the context of the retrieved information, and perform the final task.\footnote{While large pre-trained neural models have been shown to incorporate real-world knowledge in their parameters and thus may skip retrieval~\citep{petroni2019language}, they still have limited capacity and suffer from a lack of explainability.}




The standard retrieval component in many systems~\cite[e.g.,][]{fever,r3,drqa} has long relied on  term-matching methods, such as tf-idf or BM25~\citep{robertson2009probabilistic}. 
These methods rely on efficient algorithms and usually perform reasonably well regardless of the problem.
In contrast, recent neural retrieval models, such as ICT \citep{ict}, DPR \citep{dpr} and RAG \citep{rag} achieve better results by learning directly from task-specific training data and going beyond simple keyword matching.
While task specialisation results in improved task performance, researchers have observed that a retriever trained for one specific domain will typically achieve low out-of-domain performance, and even lower performance on entirely different tasks~\citep{kilt}.
This has two implications. 
First, unlike tf-idf or BM25, neural retrieval models are unsuitable for low data regimes such as few- and zero-shot settings. 
Second, task-specific retrievers complicate practical applications where multiple knowledge-intensive tasks may need to be performed using the same supporting database or over the same input text. It may not be practical to deploy multiple separate specialised models due to computational performance or memory concerns.


We ask the following question in this work: can we develop a \emph{universal} neural retriever? Namely, we target a retriever that can perform well on a wide variety of problems, \emph{without} task-specific fine-tuning, but, if additional in-domain labelled data is available, it can be further fine-tuned to improve the performance. 
%
%
We perform a large experimental study to attempt to build such a universal retrieval model. 
We find that, by jointly training on an extensive selection of retrieval tasks, we obtain a model which is not only more robust than previous approaches, but also can lead to better performance on the downstream knowledge-intensive tasks when plugged into an existing system.
Our approach combines the benefits from IR-based models with those of task-specific neural retrievers -- namely, good performance when no (or not enough) training data is available and high task performance due to its ability to learn highly specialised representations.

Our contributions can be summarised as follows.
\begin{itemize}
    \item We propose a single general-purpose ``universal'' retrieval model, able to perform comparably or better than specialised retriever approaches in both zero-shot (leave-one-out) and few-shot retrieval. We investigate several model variants, shedding light on what are the aspects of the architecture that affect its performance.
    \item We show that our model's gains in terms of retrieval directly translate into performance gains for a variety of downstream knowledge-intensive tasks.
    \item We will share the implementation as well as our best model. This is in the form of a readily available BERT checkpoint which, as we will show, can be used by NLP practitioners as a strong out-of-the-box retrieval system, but which can also undergo further in-domain training for even higher performance.
\end{itemize}



\section{Background}
\label{sec:background}

In this section, we first give an overview of retrieval methods based on sparse and dense representations. 
We then discuss a wide range of knowledge-intensive NLP tasks, where retrieval plays a crucial role in solving the problems.


\subsection{Retrieval methods}
\label{ssub:retrievalbg}

%
Given a large collection of unstructured text passages, information retrieval (IR) can be broadly defined as finding a small set of passages that satisfies an information need, often presented in the form of a short-text query~\cite{manning2008introduction}. 
Traditional IR methods, such as tf-idf and BM25~\cite{robertson2009probabilistic}, match keywords efficiently with an inverted index.
Such methods can be seen as representing queries and passages in high-dimensional, \emph{sparse} vectors, where each dimension corresponds to a term in the vocabulary and the weight indicates its importance.

%
%
In contrast to tf-idf and BM25, dense retrieval methods encode text as a latent semantic vector of a fixed, much smaller dimensionality.
Whether a passage is relevant to a given query is determined by the distance of their vectors \cite{deerwester1990indexing}.
Although dense representations do not encode tokens explicitly and can potentially map paraphrases of completely different tokens to close vectors, performance of early dense retrieval methods was often inferior to term-matching approaches, except when large labelled data is available~\cite{yih2011learning,gao2011clickthrough,Huang-DSSM13}.
%
%
Thanks to success of large pre-trained models~\cite{bert,liu2019roberta}, however, recent dense retrieval methods have shown to outperform the sparse counterparts, when fine-tuned on a small set of in-domain labelled data~\cite{dpr,rag,xiong2020approximate}. Efficient index and search of dense vectors are made possible by maximum inner product search (MIPS) algorithms~\citep[e.g.,][]{NIPS2014_5329, guo2016quantization}, as well as tools like FAISS~\cite{faiss}.

Our work is built upon the Dense Passage Retriever (DPR) architecture of \citet{dpr}, which was initially proposed for the task of open-domain question answering. DPR is a neural bi-encoder model which embeds queries with an encoder $\boldsymbol{f}\,(\cdot)$ and passages with a separate encoder $\boldsymbol{g}\,(\cdot)$. Given an input query $x$ and a target passage $y$, we have
\[
\operatorname{p}\left( x \mid y\right) \propto \operatorname{sim}(x, y),
\]
where the similarity score $\operatorname{sim}\left(x, y\right)$ is defined as the inner product of the embeddings of its arguments, $\boldsymbol{f}(x) \cdot \boldsymbol{g}(y)$. Given a query at inference time, calculating its similarity with every possible passage would be prohibitive for large knowledge sources. Therefore, DPR makes use of the FAISS library \citep{faiss} to perform fast approximate nearest neighbour search in sub-linear time. 

Training is based on a contrastive loss. Given a query $x$, a relevant passage $y$, and a set of $n$ irrelevant passages $y^-_i$, we train the model by optimising the following negative log likelihood:
\begin{small}
\[
    \mathcal{L} = -\log \frac{\exp(\operatorname{sim}(x, y)) }{\exp(\operatorname{sim}(x, y)) + \sum_{i=1}^n\exp(\operatorname{sim}\left(x, y^-_i\right)) }.
\]
\end{small}

\begin{figure}[t]
\centering
\includegraphics[width=0.5\textwidth]{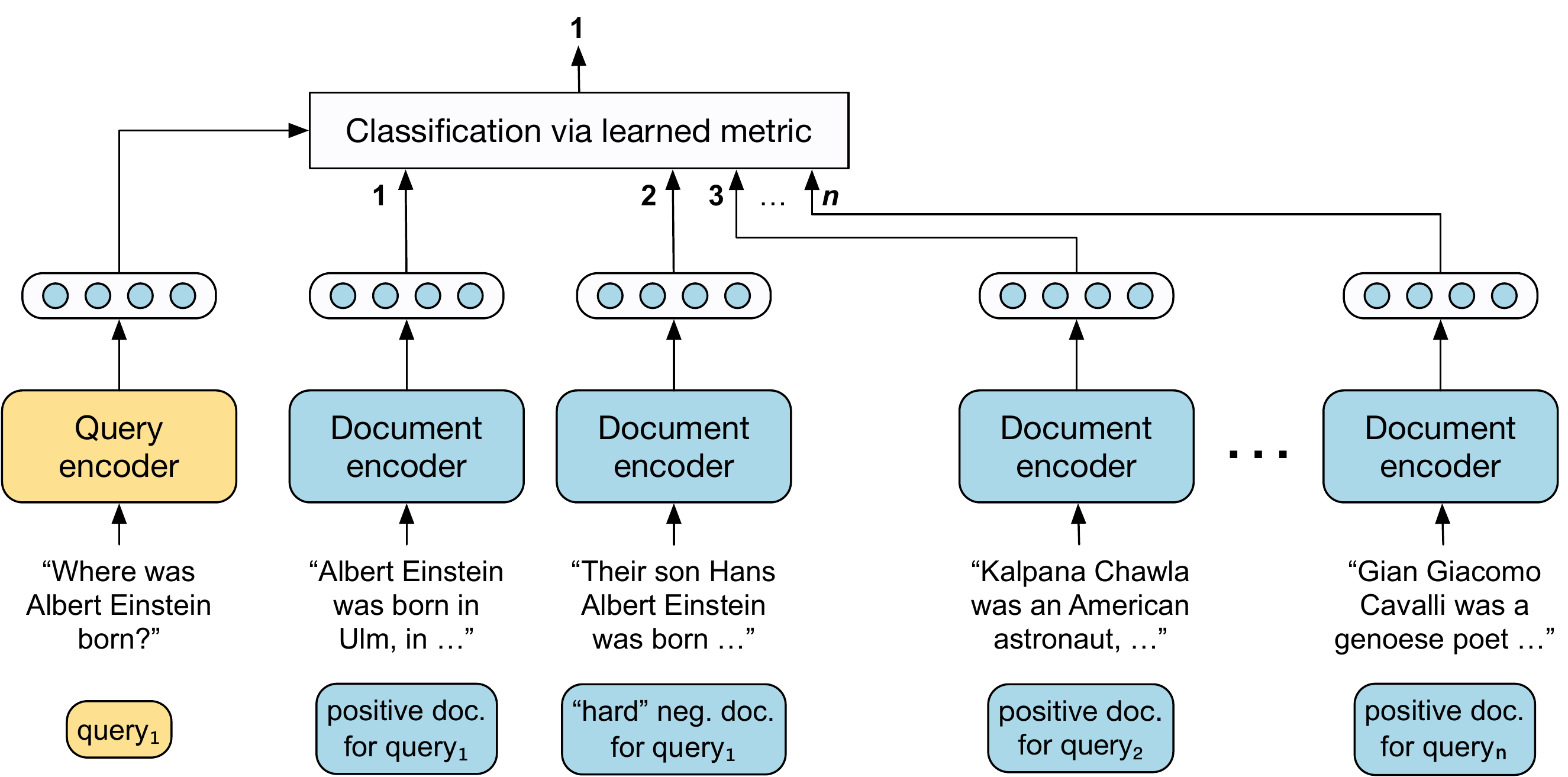}
\caption{Training of DPR \citep{dpr}, a bi-encoder model for open-domain question answering. Queries and passages are encoded as vectors, and retrieval is performed as a maximum inner product search.}\label{fig:dpr}
\end{figure}

As the set of irrelevant passages, we use the relevant passages for other queries within the same batch, as well as a specially selected ``hard'' confounder. This is a passage which has high lexical overlap with the query (high BM25 score), but is not among the set of relevant passages for the given data point. \citet{dpr} have shown that the inclusion of such ``hard'' confounders leads to substantially improved training results. This training process is illustrated in Figure \ref{fig:dpr}.

\subsection{Knowledge-intensive Tasks}

For the training and evaluation of all models in the paper we make use of KILT, a benchmark and library of datasets \citep{kilt}. KILT consists of a selection of datasets spanning five varied classes of knowledge-intensive tasks (i.e., question answering, slot filling, fact checking, dialogue, entity linking), with the aim to cover many different ways of seeking knowledge. Input queries can vary wildly from one task to the other, and include classic examples of open-domain retrieval tasks such as natural language questions and claims to be verified, as well as more unusual examples like conversation fragments and long chunks of annotated text. Crucially, all datasets distributed in KILT have been re-aligned such that they are all grounded in the same snapshot of Wikipedia, which the authors distribute. The knowledge required to answer any of the queries in the library of tasks can thus be found within the same unified knowledge source.

\begin{table*}[ht!]
\centering
\resizebox{\textwidth}{!}{   
\fontsize{8.4}{10.1}\selectfont \setlength{\tabcolsep}{0.5em}
\begin{tabular}{lp{4.6cm}p{3cm}p{4.6cm}}
 \toprule
 \textbf{Task} & \textbf{Example query} & \textbf{Answer} & \textbf{Relevant doc.} \\
 \midrule
 Question Answering & Who is playing the Halftime Show at Super Bowl 2016? & Coldplay & The Super Bowl 50 Halftime Show took place on February 7, 2016 ... It was headlined by the British rock group Coldplay.\\
 Fact Checking & Bermuda Triangle is in the western part of the Himalayas & \texttt{REFUTES} & The Bermuda Triangle ... is a loosely defined region in the western part of the North Atlantic Ocean \\
 Slot Filling & Piner Creek \texttt{[sep]} mouth of the watercourse & Santa Rosa Creek & Piner Creek discharges to Santa Rosa Creek which in turn ... \\
 Entity Linking & Leicestershire take over at top after innings victory. London. \texttt{[start\_ent]}West Indian\texttt{[end\_ent]} all-rounder Phil Simmons ... & West Indies cricket team & The West Indies cricket team is a multi-national men's cricket team representing the Anglophone Caribbean region\\
 Dialogue & I am a big fan of Star Trek \texttt{[sep]} I don't know much about it. When did the first episode air? \texttt{[sep]} It debuted in .. \texttt{[sep]} What is the plot of the show? & William Shatner plays the role of Captain Kirk & It followed the interstellar adventures of Captain James T. Kirk (William Shatner) and his crew ... \\
 \bottomrule
\end{tabular}
}
\caption{Illustrative examples of some of the tasks within KILT, and how varied their query formulations can be.}
\label{tab:kiltex}
\end{table*}

To illustrate the variety of ways in which the input queries for different tasks can be formulated, we provide a few simple examples in Table \ref{tab:kiltex}. In spite of the differences between query formulations, all these tasks share one crucial aspect: they all require a retriever to fetch the relevant passages from the knowledge source, in order to support the final downstream task.

\section{Methods}
\label{sec:methods}

\subsection{Universal retrieval}
\label{ssec:ur}

\begin{figure}[h]
\centering
\includegraphics[width=0.5\textwidth]{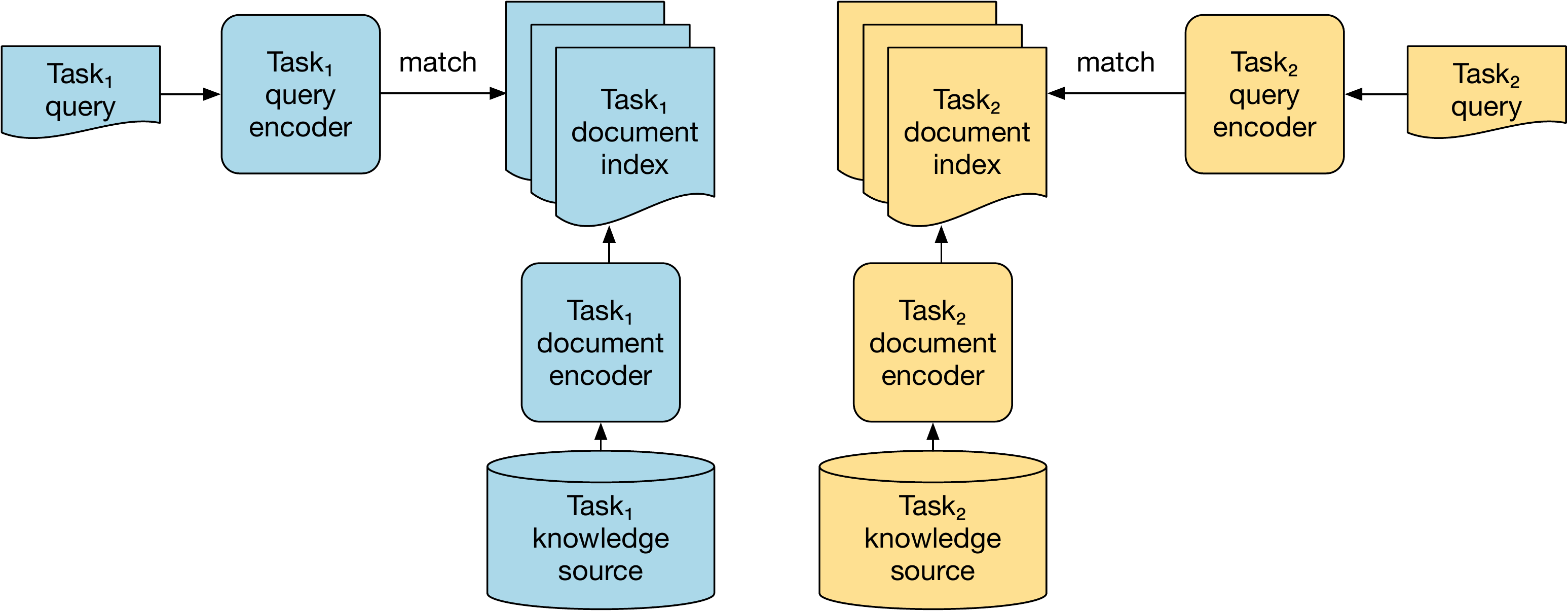}
\caption{Two retrieval tasks performed by two fully-specialised models.}\label{fig:ur-single-task}
\end{figure}

Using task-specific models to tackle our collection of retrieval tasks would involve completely separate models, one per dataset. 
As illustrated in~Figure \ref{fig:ur-single-task}, this would lead to a proliferation of models and data, down to separate indexed copies of the knowledge source itself (Wikipedia). This setup will form one of our baselines.

Multi-task training has been successfully used to allow models to leverage cross-task data, as well as to provide a regularisation effect leading to better generalisation ability \citep{mtdnn}. We apply this concept to neural retrievers, with the aim of improving performance by jointly leveraging multiple different retrieval datasets.

\begin{figure}[h]
\centering
\begin{subfigure}{.22\textwidth}
  \centering
  \includegraphics[width=.99\linewidth]{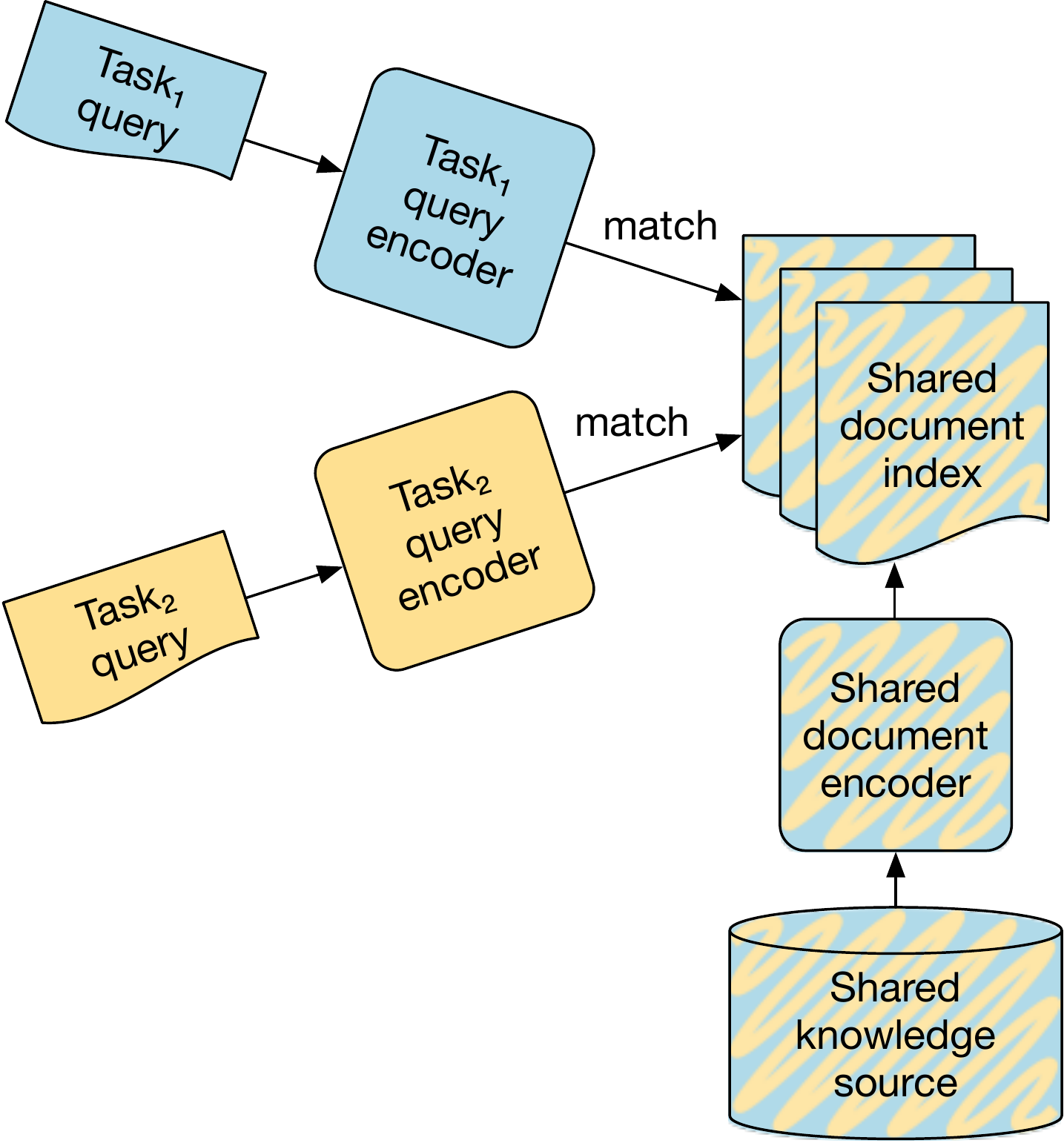}
  \caption{Separate query encoders.}
  \label{fig:ur-task-encoder}
\end{subfigure}\quad\begin{subfigure}{.22\textwidth}
  \centering\vspace{.55cm}
  \includegraphics[width=.99\linewidth]{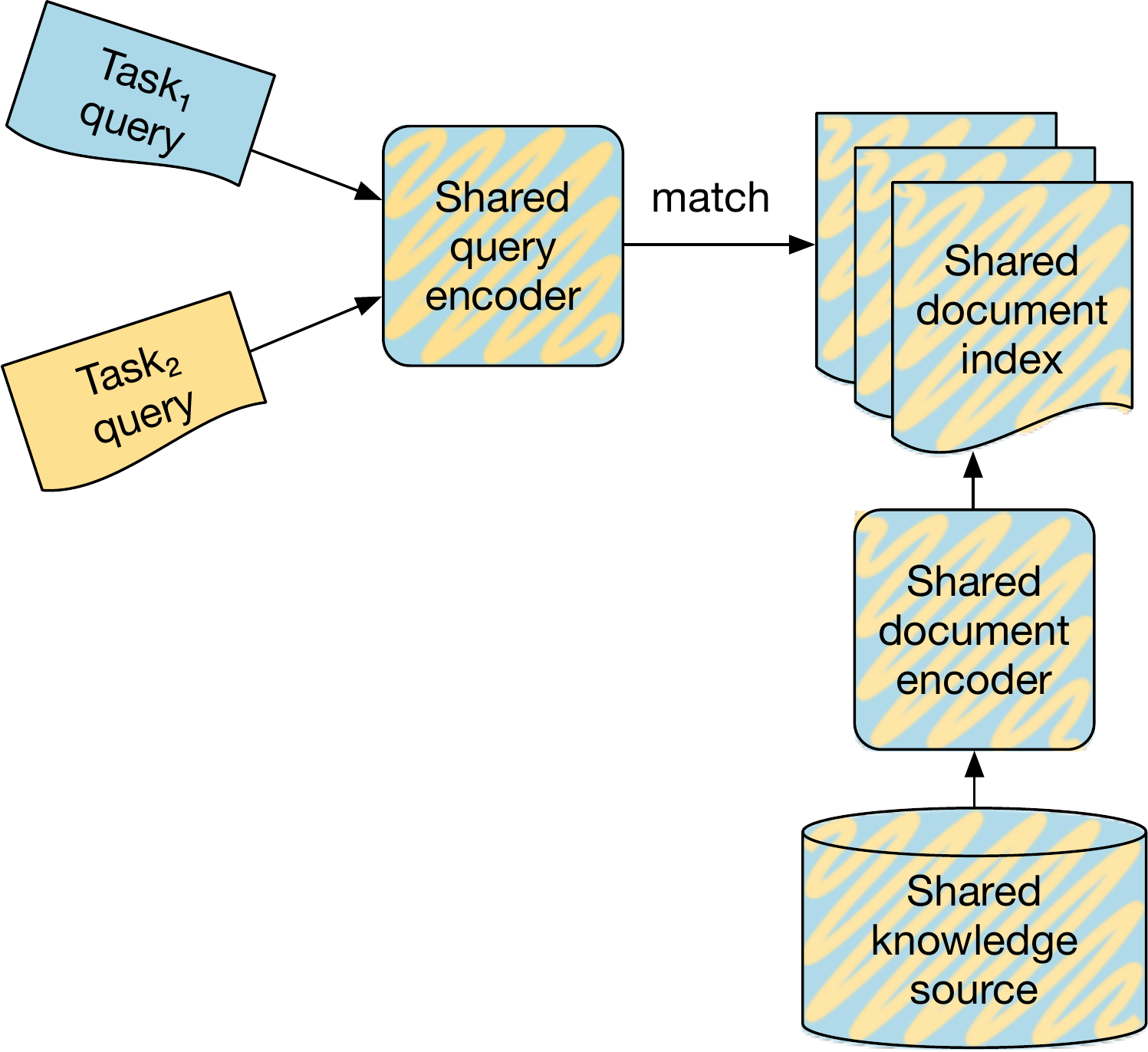}
  \caption{A single retrieval model.}
  \label{fig:ur-all-shared}
\end{subfigure}
\caption{Parameter sharing between neural retrievers.}
\label{fig:ur}
\end{figure}

Our base setup is illustrated in Figure \ref{fig:ur-all-shared} and involves using a shared passage encoder --- so that a single index of encoded passages can be used --- as well as a query encoder that is shared across all tasks. In essence, in this setup a single DPR model is used to perform all retrieval tasks.

Due to the complexity of training and evaluating retrieval models (which involves training the retriever, embedding all of Wikipedia, and building an index), our main set of experiments is all based on this configuration, which was found to work well in preliminary experiments. However, in order to report on the performance of alternative architectures, we also investigate the following additional variants in a restricted experimental setting, limited to a few tasks:
\begin{itemize}
    \item \textbf{Task-specific query encoder}. A different query encoder is used for each family of tasks, e.g. all question answering tasks use the same query encoder, but fact checking uses a different one. This is meant to allow for potentially different needs in processing queries, given the fundamentally diverse nature of the tasks at hand. This setup configuration is illustrated in Figure \ref{fig:ur-task-encoder}.
    \item \textbf{Task markers}. This approach is similar to our base setup, where a single model performs all tasks. Additionally, we introduce specialised tokens which are inserted at the beginning of each query. Their aim is to help the model distinguish between the different tasks, by marking them. We use one task marker for each of the five task classes of KILT, such that all question answering tasks share the same marker.
\end{itemize}

\subsection{Adversarial confounder selection}
\label{ssec:adv-methods}

We saw in \S~\ref{ssub:retrievalbg} how ``hard'' confounder passages are collected using a BM25 baseline, following the standard approach in DPR. However, any other retriever can be used to select such confounders, including the very retriever being trained, leading to an iterative, self-adversarial training. Concretely, this amounts to following steps: (1) a first version of the retriever is trained with BM25 confounders; (2) new confounders are selected with the trained model, by retrieving high-ranking passages which are not among the set of relevant ones; (3) a second version of the model is trained using the additional new confounders.

Intuitively, it is expected that this approach should lead to higher quality confounders compared to those selected by BM25 based on simple keyword matching. Based on our own experience as well as relevant literature \citep{colbertqa}, this adversarial approach has been shown to work well for question answering. 

As a way of further pushing the performance of the model, we experiment with this adversarial confounder selection on two datasets, Natural Questions \citep{nq} and TriviaQA \citep{trivia}. We selected these two datasets since, out of all of the tasks we are considering, they have an easy way of checking whether a certain passage is relevant or not for a given query -- namely, by checking whether the answer is present in the passage. This enabled us to automatically build sets of confounders, ensuring relevant passages would be excluded.\footnote{Strictly speaking, assuming a passage to be irrelevant because of the absence of the answer span is not formally correct. However, experiments show a good correlation between this simple check and the overall model quality.}
\section{Experiments}

\subsection{Experimental settings}

\paragraph{Dataset selection} For our experiments we select the eight KILT datasets listed in Table \ref{tab:datasets}, which cover all five task classes and include a training split, a validation split, and a held-out test split.

\begin{table}[t]
    \centering
    \begin{tabular}{llr}
        \toprule
        \textbf{Dataset} & \textbf{Task class} & \textbf{\#Train}\\
        \midrule
        FEVER & Fact Checking & 71\,k \\
        AIDA-YAGO 2 & Entity Linking & 18\,k \\
        T-REx & Slot Filling & 2,284\,k \\
        Zero Shot RE & Slot Filling & 132\,k \\
        Natural Questions & QA & 77\,k \\
        HotpotQA & QA & 69\,k \\
        TriviaQA & QA & 53\,k  \\
        Wizard of Wikipedia & Dialogue & 80\,k \\
        \bottomrule
    \end{tabular}
    \caption{KILT datasets used in this work, and the size of our converted training sets for each.}\label{tab:datasets}
\end{table}

\paragraph{Preprocessing} Starting from the raw KILT data, we split each Wikipedia article into disjoint 100-token chunks which form our basic retrieval units, following the approach of \citet{multi-passage-bert} and \citet{dpr}. To maintain the same language introduced in \S\ref{sec:methods}, we will simply call these chunks \emph{passages}.

This preprocessing results in a knowledge source of 36 million passages. In order to harmonise all datasets to the same knowledge source, KILT used a mapping strategy based on the BLEU metric to map relevant passages in the original versions of its datasets to passages in its own shared knowledge source \citep{kilt}. Entries included in the KILT training sets which have a mapping BLEU score below 0.5 are likely to be noise, and we exclude them from training.

\paragraph{Multi-tasking} Training is performed on the union of all training sets. Since two of the training sets are of different orders of magnitude, we use a simple downsampling strategy to bring them to the same order of magnitude as the others. Preliminary experiments with more complex sampling methods, like resampling all datasets so that each epoch would see an equal number of samples from each, found that they had no measurable effect compared to this simpler approach.

\paragraph{Encoders} Our query and passage encoders are initialised as two distinct BERT base uncased encoders \citep{bert}, trained separately. As pooling mechanism we find it effective to simply take the \texttt{[CLS]} token representation at the topmost layer.

\paragraph{Training} We train our models for up to 80 epochs. To select the best checkpoint, we perform full evaluations of the validation set retrieval performance at regular intervals. We use the Adam optimiser \citep{adam} with a learning rate of $2\cdot10^{-5}$ with warmup and a linear decay schedule, and a dropout rate of $0.1$. The batch size is set to $128$ samples, and in preliminary experiments we found no benefit in increasing this further. We use an additional ``hard'' confounder per batch, selected based on BM25 score as in \cite{dpr}.

\paragraph{Downstream evaluation} When evaluating our retriever within a larger architecture to perform a knowledge-intensive task, we replicate the DPR + BART setup of \citet{kilt}. This uses DPR to retrieve and prepend the top 3 passages to the query, which is then processed by a task-specific fine-tuned BART model to generate the final answer for the end task.

\subsection{Universal retrieval}
\label{ssec:general}
\begin{table*}[t]
\centering

\resizebox{\textwidth}{!}{   
\fontsize{8.4}{10.1}\selectfont \setlength{\tabcolsep}{0.5em}
\begin{tabular}{l@{\hskip 2em}rrrrrrrr}
 \toprule
   & \multicolumn{1}{r}{Fact Check.}  & \multicolumn{1}{c}{Ent. L.} & \multicolumn{2}{c}{Slot Filling} & \multicolumn{3}{c}{Open Domain QA} & \multicolumn{1}{r}{Dial.}  \\
 model  & \textbf{FEV}  & \textbf{AY2} & \textbf{T-REx} & \textbf{zsRE}  & \textbf{NQ} & \textbf{HoPo} & \textbf{TQA} & \textbf{WoW}   \\
\midrule
Multi-task & \textbf{74.72}\,/\,\textbf{46.96} & \textbf{83.78}  & \textbf{69.18}\,/\,53.54 & \underline{77.23}\,/\,\underline{41.70} & \underline{61.51}\,/\,\textbf{28.80} & \underline{44.21}\,/\,\underline{38.42} & \underline{61.95}\,/\,\underline{24.56} & 39.70\,/\,\textbf{24.07} \\
BM25 & 50.13\,/\,40.06 & 3.47 & 58.60\,/\,51.64 & 66.43\,/\,52.98 & 25.83\,/\,14.20 & 43.95\,/\,38.38 & 29.44\,/\,16.16 & 27.50\,/\,18.41 \\
\midrule
\multicolumn{9}{c}{Task-specific models}\\
\midrule
FEVER & \underline{73.60}\,/\,\underline{43.92} & 5.62 & 19.50\,/\,10.02 & 42.88\,/\,19.98 & 36.69\,/\,18.05 & 23.18\,/\,17.59 & 45.08\,/\,22.24 &  \textbf{41.27}\,/\,19.85 \\
AY2 & 47.36\,/\,{37.58} & \underline{81.77} & 5.52\,/\,\phantom{0}4.08  & 8.94\,/\,\phantom{0}5.50 & 10.22\,/\,\phantom{0}6.77 & 11.69\,/\,10.71 & 15.11\,/\,\phantom{0}8.47 & 17.59\,/\,13.08 \\
T-REx & 45.63\,/\,25.22 & 1.05 & \underline{69.08}\,/\,\textbf{58.54} & 71.64\,/\,40.95 & 17.10\,/\,\phantom{0}8.71 & 22.31\,/\,15.63 & 18.10\,/\,\phantom{0}8.06 & 4.02\,/\,\phantom{0}1.83 \\
zsRE & {70.10}\,/\,33.12 & 0.42 & 68.34\,/\,\underline{57.40} & \textbf{97.74}\,/\,\textbf{78.81} & 25.98\,/\,13.81 & 22.23\,/\,18.35 & 28.68\,/\,14.44 & 10.40\,/\,\phantom{0}2.09 \\
NQ & 68.16\,/\,14.81 & 1.44 & 31.78\,/\,\phantom{0}7.20 & 61.12\,/\,12.92 & \textbf{63.24}\,/\,\underline{28.13} & 29.39\,/\,11.33 & 48.39\,/\,14.42 & 30.77\,/\,11.81 \\
HoPo & 56.18\,/\,40.03 & 2.07 & 35.76\,/\,27.62 & 44.44\,/\,31.15 & 35.60\,/\,23.26 & \textbf{46.63}\,/\,\textbf{43.47} & 41.18\,/\,\textbf{29.37} & 23.51\,/\,16.02 \\
TQA & 70.06\,/\,10.68 & 4.95 & 32.22\,/\,12.52 & 60.37\,/\,17.43 & 45.01\,/\,12.97 & 32.62\,/\,13.05 & \textbf{65.12}\,/\,23.79 & \underline{41.17}\,/\,\phantom{0}8.11 \\
WoW & 59.16\,/\,42.79 & 3.11 & 20.92\,/\,18.52 & 41.14\,/\,35.26 & 33.27\,/\,22.52 & 20.36\,/\,17.66 & 39.37\,/\,23.15 & 40.32\,/\,\underline{20.73} \\
\bottomrule
\end{tabular}
}
\caption{Page- and passage-level $R$-precision on KILT validation data. For the AIDA-YAGO 2 dataset, due to the nature of the task, only page-level retrieval is defined.}
\label{tab:page-retrieval-dev}
\end{table*}

The results of the evaluations reported in \cite{kilt} show that retrievers trained for question answering have poor performance outside of their domain. We would like to understand if it is possible to design a single model which can accurately satisfy the information needs of a wide variety of knowledge-intensive tasks. In short: \emph{Can a neural retriever be universal?}

We perform a comprehensive evaluation of several models on the eight tasks of Table \ref{tab:datasets}. The setups we evaluate include eight task-specific models (one trained on each of the eight datasets), for which we measure both in-domain and out-of-domain performance, and a BM25 baseline. Additionally, we include a multi-task trained model -- as described in \S\ref{ssec:ur} -- with the hope that it can learn to perform all tasks satisfyingly. This amounts to 10 models evaluated on eight tasks each, for a total of 80 evaluations.

To measure retrieval performance, we adopt the main metric used for the KILT benchmark, $R$-precision. This is calculated as $r/R$, where $R$ is the total number of relevant passages for a given query, and $r$ is the number of relevant passages returned among the top-$R$ retrieval results. For the case of $R=1$ this is therefore equivalent to precision@1. Table~\ref{tab:page-retrieval-dev} shows retrieval performance on the validation data, with the best performance on a given dataset marked in bold, and the second best performance underlined.

While the KILT evaluation focuses on retrieval at the level of Wikipedia pages (thereby marking as ``hits'' any results that lie within the correct page), we are also interested in performing an evaluation at a more fine-grained level. We therefore also evaluate our models at the passage level, using a modified version of the official KILT evaluation scripts. These are shown as the second number in each column.

We straight away notice that the task-specific models tend to achieve high performance on their respective tasks, often taking one of the top two spots. Interestingly, we also note that these neural retrievers consistently outperform the BM25 baseline, showing that the result which \citet{dpr} achieved for open-domain question answering also holds for other knowledge-intensive tasks.

The results reveal a strong performance for the multi-task model, confirming the hypothesis that a single model can be successfully trained to perform a wide variety of retrieval tasks. With the exception of one dataset, the shared model achieves the best retrieval performance or is within a few percentage points of the top score. We note that the one exception is the Zero-shot RE task \citep{zsre}, a trivial task in which the query will always contain the title of the page to be retrieved. Indeed, the model specific to this task manages to achieve a near-perfect score.

Another task which stands out for being markedly different in formulation is AIDA-YAGO~2 \citep{ay2}. As shown in Table \ref{tab:datasets}, models that were not trained on this specific task perform it very poorly. Entity linking is a task that is normally better performed by models which are explicitly designed for it \citep{genre}. We nevertheless include it to showcase the ability of neural retrievers to adapt to it, and note how well the multi-task retriever performs on it in spite of its unusual nature.

\subsection{Downstream performance}
\label{ssec:downstream}

\begin{table*}[ht]
\centering
\resizebox{\textwidth}{!}{   
\fontsize{8.4}{10.1}\selectfont \setlength{\tabcolsep}{0.5em}
\begin{tabular}{l@{\hskip 2em}rrrrrrr@{\hskip 2em}r}
 \toprule
    & \multicolumn{1}{r}{Fact Check.}  & \multicolumn{1}{c}{Ent. L.} & \multicolumn{1}{c}{Slot Fill.} & \multicolumn{3}{c}{Open Domain QA} & \multicolumn{1}{c}{Dial.}  \\
Model & \textbf{FEV}  & \textbf{AY2} & \textbf{zsRE}  & \textbf{NQ} & \textbf{HoPo} & \textbf{TQA} & \textbf{WoW} & \textbf{Avg.}  \\
\midrule
{Multi-task} + BART & \underline{86.32} & \textbf{82.61} & \textbf{57.95} & 39.75 & \textbf{31.77} & \underline{59.60} & \underline{15.33} & \textbf{53.33} \\
\midrule
DPR + BART & \textbf{86.74} & \underline{75.49} & 30.43 & \underline{41.27} & 25.18 & {58.55} & \textbf{15.55} & 47.60\\
RAG & 86.31 & 72.62 & \underline{44.74} & \textbf{44.39} & \underline{26.97} & \textbf{71.27} & 13.22 & \underline{51.36} \\
T5 & 76.30 & 74.05 & 9.02 & 19.60 & 12.64 & 18.11 & 13.49 & 31.89 \\
\bottomrule
\end{tabular}
}
\caption{KILT \emph{test} scores on the downstream evaluation. Results in the bottom section are as reported in \citet{kilt}. The score metrics are accuracy for fact checking, entitly linking and slot filling; exact match for QA; and F1 score for dialogue.\protect\footnotemark}
\label{tab:kilt-scores}
\end{table*}

We saw that our proposed approach achieves strong performance across a variety of retrieval tasks. However, our interest in neural retrievers stems from their use as components within larger systems, to perform tasks such as question answering. Our next experimental question is therefore: \emph{Can a universal retriever lead to better downstream performance in knowledge-intensive tasks?}

We perform a downstream evaluation of our approach used in conjunction with BART \citep{bart} as the generative component or classifier, adopting the same setup as \citet{kilt}. Results are reported in Table \ref{tab:kilt-scores}, with bold and underline marking the best and second best scores respectively.

The \emph{DPR + BART} line refers to a setup similar to our own, but with the simpler retriever of \citet{dpr}. Therefore, comparing its performance to ours gives us a clear indication of the contribution of multi-task training on the overall performance on knowledge-intensive tasks. Our proposed model achieves significantly better performance than this baseline in AY2, zsRE and HoPo; while for the other tasks, the discrepancy is always below two points. This fact is reflected in the last column too, showing that on average multi-task training leads to better downstream performance. The model also compares favourably to RAG \citep{rag}, a more advanced system in which the query encoder is fine-tuned on the end task.

\footnotetext{Performing this evaluation required retrieving relevant documents for all training sets. Due to the very large size of T-REx, this particular dataset could not be included in this section.}

\subsection{Zero- and few-shot performance}
\label{ssec:fewshot}

\begin{table*}[t]
\centering

\resizebox{\textwidth}{!}{   
\fontsize{8.4}{10.1}\selectfont \setlength{\tabcolsep}{0.5em}
\begin{tabular}{lrrrrrrrrr}
 \toprule
 Model  & \textbf{FEV}  & \textbf{AY2} & \textbf{T-REx} & \textbf{zsRE}  & \textbf{NQ} & \textbf{HoPo} & \textbf{TQA} & \textbf{WoW} & \textbf{Avg.}  \\
\midrule
BM25 & 50.13\,/\,{40.06} & 3.47 & 58.60\,/\,\textbf{51.64} & 66.43\,/\,{52.98} & 25.83\,/\,14.20 & \textbf{43.95}\,/\,\textbf{38.38} & 29.44\,/\,16.16 & 27.50\,/\,{18.41} & 38.17\,/\,\underline{33.12} \\
\midrule
\multicolumn{10}{c}{Leave-one-out multi-task models}\\
\midrule
Zero-shot & \underline{74.11}\,/\,37.09 & {4.16} & \textbf{67.54}\,/\,\underline{44.84} & {73.42}\,/\,32.65 & {47.23}\,/\,\textbf{21.50} & 34.72\,/\,16.52 & {49.08}\,/\,\textbf{28.06} & {36.92}\,/\,16.19 & 48.40\,/\,28.12 \\
Finetune (128) & \textbf{75.95}\,/\,32.75 & 32.38 & \textbf{67.54}\,/\,\underline{44.84} & 73.41\,/\,32.65 & \underline{47.48}\,/\,14.98 & 34.72\,/\,27.82 & \underline{54.71}\,/\,19.82 & \textbf{48.36}\,/\,17.46 & \underline{54.23}\,/\,27.19\\
Finetune (1k) & {73.08}\,/\,\underline{40.83} & \underline{70.40} & \textbf{67.54}\,/\,\underline{44.84} & \textbf{93.04}\,/\,\underline{58.67} & \textbf{51.00}\,/\,\underline{19.90} & \underline{39.19}\,/\,\underline{35.43} & \textbf{59.08}\,/\,\underline{20.22} & \underline{47.65}\,/\,\underline{19.75} & \textbf{62.62}\,/\,\textbf{34.23}\\
\midrule
\multicolumn{10}{c}{Vanilla DPR models}\\
\midrule
Finetune (128) & 37.99\,/\,25.31 & 26.23 & 0.20\,/\,\phantom{0}0.02 & 0.16\,/\,\phantom{0}0.00 & 20.92\,/\,\phantom{0}9.52 & 14.46\,/\,14.08 & 26.85\,/\,10.54 & 30.31\,/\,17.20 & 19.64\,/\,10.95\\
Finetune (1k) & 70.87\,/\,\textbf{47.82} & \textbf{72.49} &  0.20\,/\,\phantom{0}0.02 & \underline{90.33}\,/\,\textbf{80.20} & 43.43\,/\,{19.81} & 30.75\,/\,30.50 & 52.50\,/\,17.33 & 44.70\,/\,\textbf{24.92} & 50.66\,/\,31.51\\
%
\bottomrule
\end{tabular}
}
\caption{Page- and passage-level $R$-Precision in the zero-shot setting and with additional fine-tuning of 128 and 1,024 examples. We also compare to a BM25 retriever and a DPR model initialised with BERT weights.}
\label{tab:leave-one-out}
\end{table*}

Task-specific neural retrievers can achieve higher performance than IR-based methods, but they are not suitable for cases where no training data (or not enough) is available. In those cases, tf-idf and BM25 are the better choice.  To evaluate the performance of a multi-task retriever as a suitable replacement for them in this scenario, we run a series of experiments in the low data regimes (few-shot and zero-shot).

We start by training a set of multi-task retrievers (using the base setup) in the leave-one-out setting for each of the datasets, in order to see how a neural retriever will perform when trained on all domains except for the one it is to be evaluated on. The results of these zero-shot experiments are reported in the second line of Table \ref{tab:leave-one-out} (again, text here is in bold for the best overall performance, and underlined for second best). They show that, even in the zero-shot setting, the multi-task neural retriever achieves performance that is competitive to BM25, with retrieval being 10 points higher at the page level and 5 points lower at the passage level on average.

The advantage of neural retrievers over BM25 lies in their ability to improve with training. We therefore look at few-shot training for each task, and create two smaller copies for each of the original training sets with a random sample of 128 and 1,024 examples respectively. In order to evaluate the suitability of a multi-task trained retriever as a starting checkpoint for few-shot training, we take the various leave-one-out models and fine-tune them on our few-shot training sets. To check whether multi-task pre-training is effective, we also compare these to DPR models (which are just initialised with BERT weights) fine-tuned on the same data.

The bottom two sections of Table \ref{tab:leave-one-out} report the results. The most dramatic gains from fine-tuning are seen for AY2, an ``outlier'' task whose formulation differs from that of the other tasks, and which seems to benefit the most from seeing in-domain data. The zsRE performance does not seem to improve from fine-tuning on the smaller dataset, but sees a very big jump when switching to the larger dataset. As a reminder, in this trivial task the title of the page to be retrieved always appears at the start of the query. It is therefore not surprising that models specifically fine-tuned on it can achieve near-perfect scores, as long as enough training data is provided.

In spite of the fine-tuning, we note that both DPR and the multi-task model fail to improve on their performance for T-REx, suggesting that large amounts of training data are required to learn this task. Nevertheless, the multi-task model proves itself more robust, and achieves the top performance on it.

Finally, we note for 2 out of 8 tasks, namely zsRE and WoW, DPR achieves lower page-level retrieval scores than the multi-task model, but performs better at the passage level. This shows that fine-grained and coarse-grained retrieval performance are not always perfectly correlated.

Overall, the experiments show strong results for the multi-task model, with the average zero-shot performance being competitive to BM25, and the average few-shot performance being markedly better than the alternatives. The discrepancy in performance between a vanilla DPR model and the leave-one-out multi-task model is especially noticeable when using the smaller of the two datasets, in which case average performance for the latter is more than double that of vanilla DPR.



\subsection{Model variants}
\label{ssec:variants-exp}

\begin{table}[ht]
\centering
\fontsize{8.4}{10.1}\selectfont \setlength{\tabcolsep}{0.5em}
\begin{tabular}{lrrr}
 \toprule
variant & \textbf{FEV} & \textbf{NQ} & \textbf{TQA}   \\
\midrule
Base & \textbf{76.38}\,/\,40.76 & 60.91\,/\,24.50 & \textbf{64.77}\,/\,\textbf{21.75} \\
Task markers & {75.84}\,/\,\textbf{40.79} & \textbf{62.31}\,/\,25.10 & 64.04\,/\,20.86\\
Task-spec. enc. & 73.53\,/\,40.02 & 61.05\,/\,\textbf{25.52} & 64.17\,/\,21.23\\
\bottomrule
\end{tabular}
\caption{Multi-task model variants evaluated on a subset of tasks ($R$-precision on validation data at page/passage level).}
\label{tab:variants}
\end{table}

In this set of experiments we compare our base multi-task model with the two variants described in \S~\ref{ssec:ur}. Due to the high memory consumption of the ``task-specific encoders'' variant (requiring one full query encoder per task family, in addition to the passage encoder), it was only possible to perform these evaluations in a restricted setting of three datasets. The results in Table~\ref{tab:variants} do not reveal a clear winner, suggesting that the base architecture might be the better choice due to its simplicity and generally good performance.\footnote{Not included in this table, due to very poor performance in preliminary experiments, are two further variants: a base model with a single encoder for both queries and passages, and a base model trained from scratch without BERT pre-training.} 

\subsection{Adversarial confounder selection}
\label{ssec:adv-exp}
\begin{table*}[t]
\centering

\resizebox{\textwidth}{!}{   
\fontsize{8.4}{10.1}\selectfont \setlength{\tabcolsep}{0.5em}
\begin{tabular}{lrrrrrrrr}
 \toprule
   & \multicolumn{1}{r}{Fact Check.}  & \multicolumn{1}{c}{Ent. L.} & \multicolumn{2}{c}{Slot Filling} & \multicolumn{3}{c}{Open Domain QA} & \multicolumn{1}{r}{Dial.}  \\
 confounders  & \textbf{FEV}  & \textbf{AY2} & \textbf{T-REx} & \textbf{zsRE}  & \textbf{NQ} & \textbf{HoPo} & \textbf{TQA} & \textbf{WoW}   \\
\midrule
BM25 & {74.72}\,/\,{46.96} & {83.78}  & {69.18}\,/\,53.54 & {77.23}\,/\,{41.70} & \textbf{61.51}\,/\,{28.80} & \textbf{44.21}\,/\,{38.42} & \textbf{61.95}\,/\,{24.56} & 39.70\,/\,{24.07} \\
BM25 + adv & \textbf{74.79}\,/\,\textbf{52.12} & \textbf{84.86} & \textbf{71.36}\,/\,\textbf{61.40} & \textbf{80.04}\,/\,\textbf{54.08} & 59.25\,/\,\textbf{40.11} & 44.08\,/\,\textbf{41.04} & 59.19\,/\,\textbf{34.17} & \textbf{41.04}\,/\,\textbf{24.62}\\
\bottomrule
\end{tabular}
}
\caption{Comparison of two confounder selection methods for the multi-task model: simple BM25, and BM25 augmented with adversarial confounders ($R$-precision on validation data at page/passage level).}
\label{tab:adversarial}
\end{table*}

Finally, we evaluate the adversarial confounder selection method described in \S~\ref{ssec:adv-methods}. This involves augmenting our regular training sets with additional confounders for TriviaQA and Natural Questions, selected using our top multi-task trained model. A new multi-task model is then trained from scratch on this augmented data. Its performance is reported in Table~\ref{tab:adversarial}, showing an overall improvement across multiple tasks. While this approach is demonstrated here on our multi-task model, it is in fact orthogonal to it, and could be applied to any other neural retrievers trained with a contrastive loss.
\section{Related work}

The approach most closely related to ours is DPR \citep{dpr}, upon which we built all our retrieval systems. This model is covered in detail in \S~\ref{ssub:retrievalbg}, in addition to the historical context. Another closely related approach is the Retrieval-Augmented Generation (RAG) model of \citet{rag}. In its base configuration it augments DPR with a generative reader, and it trains the query encoder end-to-end (differing from traditional retriever-reader architectures which treat the two steps as disjoint). A natural extension of the work we have presented would be to combine RAG with our joint learning approach, to study whether it can lead to further gains in performance or robustness.

A number of promising techniques to boost retrieval performance have been proposed recently. These are orthogonal to our work, and as such they could be combined with it. Amongst these, pre-training methods form one class. Inverse Cloze Task \citep{ict} and its extensions \citep{retrieval-pretraining} are self-supervised pre-training methods designed for retrieval in open-domain question answering. Whether such specific pre-training is beneficial to tasks other than question answering remains an open question. CERT \citep{cert} is an alternative pre-training approach, inspired by some recent advances in computer vision. While to our knowledge this has not been applied to retrieval problems, we believe it might be promising due to its focus on sentence-level semantics (as opposed to the more standard masked language modelling pre-training, which focuses on the token-level).

Another class of orthogonal improvements to dense retrieval involves models which embed passages into multiple fixed-size vectors. Of these, ColBERT \citep{colbert} and ME-BERT \citep{mebert} are two representative examples. One further approach is ColBERT-QA \citep{colbertqa}, which additionally uses a data augmentation strategy closely related to our own approach described in \S~\ref{ssec:adv-methods}.

Finally two entity linkers, GENRE \citep{genre} and BLINK \citep{blink}, are worth mentioning. Being trained specifically for entity linking, these models will generally outperform retrieval-based approaches on that task. While they are not comparable to retrieval models and will not generally be applicable to information retrieval tasks, we mention them here to provide readers with a fuller context of the existing literature.
\section{Conclusions}

We have conducted a large-scale experimental study on knowledge-intensive tasks, and how retrieval models that tackle them seek the required information from knowledge bases such as Wikipedia.

The study started with the question of whether the way in which information is embedded for retrieval purposes is universal. Section~\ref{ssec:general} provided evidence that to a large extent it is, with a single ``universal'' retriever, trained jointly on 8 datasets, often performing comparably to task-specific models.

Armed with this knowledge, in Section~\ref{ssec:downstream} we plugged our single model in a larger pipeline, in order to see its contribution to the downstream performance on a wide range of knowledge-intensive tasks. This led to an overall improvement in downstream performance, setting new top results for a number of tasks in the KILT benchmark.

Next, in Section~\ref{ssec:fewshot}, we evaluated the model's performance in the zero-shot and few-shot settings. By evaluating on a wide range of tasks, we were able to show that our proposed approach performs comparably to BM25 in the zero shot setting, and quickly overtakes it even with minimal in-domain training.

In Section~\ref{ssec:variants-exp} we evaluated a number of more complex variants of the model involving task specialisation, but failed to see clear performance improvements. Finally, in Section~\ref{ssec:adv-exp} we saw how a simple iterative approach to data augmentation can lead to better performance.

In the coming months we will provide a pretrained snapshot of our best-performing model, in the form of a BERT checkpoint. As shown, this model will be useful in zero-shot and few-shot settings as a better performing alternative to both IR-based approaches such as BM25, as well as task-specific models. The multi-task training approach demonstrated here can also be useful in industry settings where several retrieval operations may need to be performed on the same piece of content,\footnote{E.g. fact checking and hate speech detection.} and the deployment of multiple task-specific models might not be possible due to space or computational performance concerns.

\bibliography{acl2020}
\bibliographystyle{acl_natbib}

\end{document}